\documentclass{article}

\usepackage[preprint]{corl_2026} 
\usepackage{corl_2026}  
\usepackage{amsmath}
\usepackage{amssymb}
\usepackage{booktabs}
\usepackage{graphicx}
\usepackage{placeins}
\usepackage{enumitem}
\usepackage{subcaption}

\title{Act on What You See: Unlocking Safe Social Navigation in Vision-Language-Action Models}

\usepackage{amssymb}
\usepackage{booktabs}
\usepackage{graphicx}
\usepackage{placeins}
\usepackage{enumitem}
\usepackage{subcaption}
\title{Act on What You See: Unlocking Safe Social Navigation in Vision-Language-Action Models}
\author{
  Qingzi Wang\\
  University of Maryland \\
  \texttt{qwang812@umd.edu} \\
  \And
  Xiyang Wu\\
  University of Maryland\\
  \texttt{wuxiyang@umd.edu}
  \And
  Guangyao Shi\\
  University of Southern California\\
  \texttt{shig@usc.edu}
  \And
  Dianwei Chen\\
  University of Maryland\\
  \texttt{dwchen98@umd.edu} \\
  \And
  Xianfeng Yang\\
  University of Maryland\\
  \texttt{xtyang@umd.edu}
  \And
  Dinesh Manocha \\
  University of Maryland \\
  \texttt{dmanocha@umd.edu} \\
}

\newcommand{\wxy}[1]{\textbf{\textcolor{blue}{$\leftarrow$ #1}}}
\newcommand{\sgy}[1]{\textbf{\textcolor{magenta}{$\leftarrow$ #1}}}
\newcommand{\ours}{SALSA}

\graphicspath{{figures/}}
\begin{document}
\maketitle

\begin{abstract}
    Safe social navigation requires robots to distinguish people from ordinary 
obstacles and to react before danger becomes imminent. We show that pretrained 
Vision-Language-Action (VLA) models already encode pedestrian-object distinctions 
and future collision signals in their internal representations, but behavior cloning 
fails to translate these signals into socially appropriate actions. To address this mismatch, we propose SALSA, a two-stage annotation-free post-training 
framework: (1) social behavioral alignment bridges intermediate-layer social features 
to the action head and trains on counterfactual human--object scene pairs to break 
visual saliency shortcuts; (2) temporal safety alignment provides automatically 
generated future-risk supervision to enable anticipatory collision avoidance. 
On SCAND and real-world deployment, SALSA reduces near-collisions by 86.4\% 
and improves social counterfactual accuracy from 53\% to 93\%, demonstrating that 
safer social navigation can be achieved by teaching VLA policies to act on 
representations they already possess.
    These results show that pretrained VLA policies can be adapted for safer social navigation by better aligning their latent representations with action generation.
\end{abstract}



\section{Introduction}
\label{sec:introduction}

Social navigation asks robots to do more than avoid obstacles: they must move around people safely, legibly, and appropriately~\cite{francis2025principles, singamaneni2024survey}. 
%
%
The same spatial layout can therefore call for different actions depending on who the robot is interacting with.~\cite{cui2021learning, nguyen2023toward, seneviratne2026chop, fang2026obstacles}.
%
%
Recent navigation Vision-Language-Action models (VLAs) offer a path toward general-purpose policies by using large-scale behavior cloning to map visual observations and goals directly to robot actions~\cite{cheng2024navila,payandeh2024social, hirose2025omnivla, huang2026tic}.
However, behavior cloning optimizes action reproduction, not social decision making~\cite{han2025ratatouille}.
Even when pretrained vision-language backbones encode useful information about people and context~\cite{chen2024vision}, the learned policy may fail to turn it into appropriate behavior.
%
%
This mismatch between perception and behavior has become a key limitation of current VLAs in social navigation.~\cite{francis2025principles,song2024vlmsocialnavsociallyawarerobot, zhang2025safevla}.

A straightforward way to address this limitation is to introduce manually annotated social supervision or auxiliary reasoning modules specifically designed for social navigation~\cite{kong2025autospatial, elnoor2025vi}. However, such approaches undermine the scalability and end-to-end simplicity that make VLAs attractive as a general navigation solution. 
We therefore investigate whether the limitation truly stems from a lack of social scene understanding, or from a failure to translate existing internal representations into action. 



\noindent {\bf Main Results:} Our key finding is that the bottleneck in current navigation VLAs may not lie in scene understanding itself, but in how this understanding is translated into behavior. 
Through probing analysis, we find that navigation VLAs already encode signals related to social distinction and future collision risk in their internal representations. However, these signals fail to influence behavior.

 Motivated by this observation, we propose \ours{}, a two-stage annotation-free post-training approach for navigation VLA models. 
Rather than adding human social annotations or inference-time reasoning modules, \ours{} improves how the policy uses its existing internal representations. 
The first stage, \textit{social behavioral alignment}, aligns action generation with human--object counterfactual scene pairs, encouraging the policy to produce different behaviors in geometrically similar but socially different situations. 
The second stage, \textit{temporal safety alignment} automatically relabels trajectories with future-risk signals, enabling anticipatory responses before collisions become imminent. 
Together, these stages convert latent social and safety-related representations into behavior-level supervision, improving social navigation while preserving the model's unified end-to-end inference structure.
Our main contributions include:

\begin{itemize}[leftmargin=*, itemsep=2pt, topsep=2pt]
    \item We systematically reveal the perception--behavior mismatch in navigation VLAs: social and risk-related information is encoded internally, yet fails to effectively influence behavioral decisions.
    \item We propose a unified post-training framework that leverages human--object counterfactual scene pairs and temporal relabeling-based supervision, encouraging socially distinct navigation while enabling anticipatory responses to potential collisions without additional human annotation.
    \item We evaluate our framework on SCAND and in real-world robot deployment. Compared with the original behavioral-cloning VLA, our method reduces near-collision rates by 86.4\%, improves social counterfactual accuracy from 53\% to 93\%, and increases real-world pedestrian clearance by 76\%, demonstrating safer and more socially sensitive navigation.
\end{itemize}

\section{Related Work}
\label{sec:related work}

\textbf{Socially Aware Visual Navigation.} 
Social navigation~\cite{francis2025principles} asks robots to move safely in human-populated spaces while accounting for social context. Classical approaches such as the Social Force Model~\cite{helbing1995social} model pedestrian comfort with repulsive potentials, while learning-based methods add social awareness through rewards, representations, or planning objectives~\cite{chen2016decentralizednoncommunicatingmultiagentcollision,chen2018sociallyawaremotionplanning, chen2019crowdrobotinteractioncrowdawarerobot, wu2023intent,payandeh2024social}. These systems often rely on modular representations such as pedestrian detection and tracking.
Recent work uses vision-language models for higher-level social reasoning. Path-Etiquette~\cite{fang2026obstaclesetiquetterobotsocial}, VLM-Social-Nav~\cite{song2024vlmsocialnavsociallyawarerobot}, and GSON~\cite{luo2025gsongroupbasedsocialnavigation} use VLMs to rank paths, score behaviors, or reason about social groups, but remain modular. Related systems such as ZSORN~\cite{guan2025zsorn} and Social-LLaVA~\cite{payandeh2024social} improve object-centric navigation and explainable social reasoning, yet do not provide an end-to-end visuomotor account of socially differentiated behavior.
End-to-end visual navigation models such as GNM~\cite{shah2023gnmgeneralnavigationmodel} and ViNT~\cite{shah2023vintfoundationmodelvisual} learn strong navigation priors from behavior-cloned demonstrations, but are limited to behaviors present in the data. SELFI~\cite{hirose2024selfiautonomousselfimprovementreinforcement} studies online self-improvement, while SocialNav-SUB~\cite{munje2025socialnavsubbenchmarkingvlmsscene} shows that current VLMs still lag behind simple rules and human-consensus baselines on social-scene VQA. CHOP~\cite{seneviratne2026chop} uses counterfactual preference labels to improve safety and goal completion, but the internal basis for socially differentiated navigation remains unclear. 
%
%
This motivates our representation-level analysis.

\textbf{Probing Social Representations in Navigation VLAs.}
Explaining VLA behavior requires looking beyond the final action: representation analysis can show whether a model encodes social navigation cues, and whether failures come from missing concepts or poor translation into control. 
\citet{chefer2021generic}, \citet{aflalo2022vl}, and CLIP-Dissect~\cite{oikarinen2022clip} provide tools for identifying visual and linguistic concepts inside multimodal representations, rather than only judging outputs.
Navigation VLA systems make this question concrete. NaVILA~\cite{cheng2024navila} and VAMOS~\cite{castro2025vamos} separate semantic planning from low-level embodiment, so the planner can be probed for humans, blocked paths, social constraints, or yielding decisions.
Recent VLA interpretability work extends this idea beyond visualization: \citet{buurmeijer2026observingcontrollingfeaturesvisionlanguageaction} decode and control manipulation features from hidden states, \citet{lu2025probingvisionlanguageactionmodelsymbolic} recover symbolic states from OpenVLA layers, and \citet{häon2025mechanisticinterpretabilitysteeringvisionlanguageaction} steer behavior through semantic activation directions. Although focused on manipulation, these tools can test whether navigation VLAs encode pedestrian intent, personal-space violations, group structure, or stop/yield choices.
Building on these tools, we probe navigation VLAs for social and temporal cues that should guide safe behavior. We find that these cues appear in their internal representations, but behavior cloning does not reliably turn them into socially appropriate actions. This motivates targeted training that transfers latent social understanding and risk awareness into end-to-end navigation.

 \section{Preliminary: Behavioral Attribution of VLA Navigation Decisions} 
\label{sec:behavioral_attribution}

Before designing an alignment method, we first diagnose why a behavior-cloned navigation VLA fails to exhibit socially appropriate navigation. We ask whether the failure comes from missing perception that the model never distinguishes humans from objects or safe from dangerous situations, or from a failure to translate such internal information into action. To answer this, we combine behavioral counterfactual tests with layer-wise representation probing.

\vspace{-7pt}
\subsection{What Drives VLA Avoidance Behavior?}
\label{sec:what_drives}
\textbf{Experimental Setup.}
We select pedestrian scenes from the SCAND dataset~\cite{karnan2022socially}, replace pedestrians with static objects via inpainting, and compare the base VLA's predicted trajectories across matched conditions. Because the scene geometry is preserved, changes in trajectory can be attributed to visual or semantic differences rather than to changes in obstacle location.

\textbf{Test 1: Visual Saliency Dominates Avoidance Behavior.}
We first test whether avoidance magnitude is explained by visual saliency, measured as the pixel-level contrast between the obstacle region and the surrounding background. Across human and object counterfactuals, avoidance magnitude correlates strongly with saliency. To isolate this effect, we generate low-, original-, and high-contrast variants of the same pedestrian at the same position. The model detours farther as the pedestrian becomes more visually prominent, indicating that avoidance strength is driven primarily by appearance contrast rather than social meaning.

\textbf{Test 2: No Systematic Social Differentiation.}
We then test whether the model reacts differently to humans and objects when scene geometry is matched. Across speed adjustment, lateral distance, detour direction, and path selection, the base VLA shows no systematic differentiation between pedestrians and static objects.

These behavioral counterfactuals suggest that the VLA's avoidance behavior is governed by low-level visual saliency rather than social semantics. However, this does not yet reveal whether the model lacks social perception entirely or whether it encodes social information internally but fails to use it for control. We therefore probe the model's hidden representations.

\vspace{-5pt}
\subsection{Internal Representation Analysis}
\label{sec:representation_analysis}

\textbf{Social Entity Signal.}
Linear probes trained on counterfactual human-object scene pairs show that social distinction is most readily decodable at intermediate layers (peaking around Layer 9) and becomes progressively less linearly salient toward the output layer, where it is encoded in a more distributed, lower-margin form. This indicates that social information, while present throughout the network, is most accessible in intermediate representations and harder to extract at the output layer that feeds the action head.

\textbf{Temporal Danger Signal.}
Using automatically generated labels based on future collision outcomes, linear probes decode approaching danger at roughly 74\% accuracy. In contrast to the social signal, this decodability is essentially uniform across depth and persists undiminished at the output layer that feeds the action head. Yet, as the behavioral analysis shows, this readily available signal triggers no preventive action.

Together, these results reveal a representation-behavior gap: the base model encodes both social entities and impending danger, yet neither shapes its actions. In both cases the signal is decodable at the output layer that feeds the action head, so it is unused, not absent. The two differ in how the signal is distributed, and thus in how we address it. Social decodability peaks at intermediate layers, so we route the more salient mid-layer representation to the action head and train it with counterfactual contrastive supervision. Danger decodability is uniform across depth, so we instead supply the missing action-level link to preventive behavior through temporal hindsight supervision. These two interventions form our two-stage framework.

\section{Method}
\begin{figure*}[t]
\centering
\includegraphics[width=\textwidth]{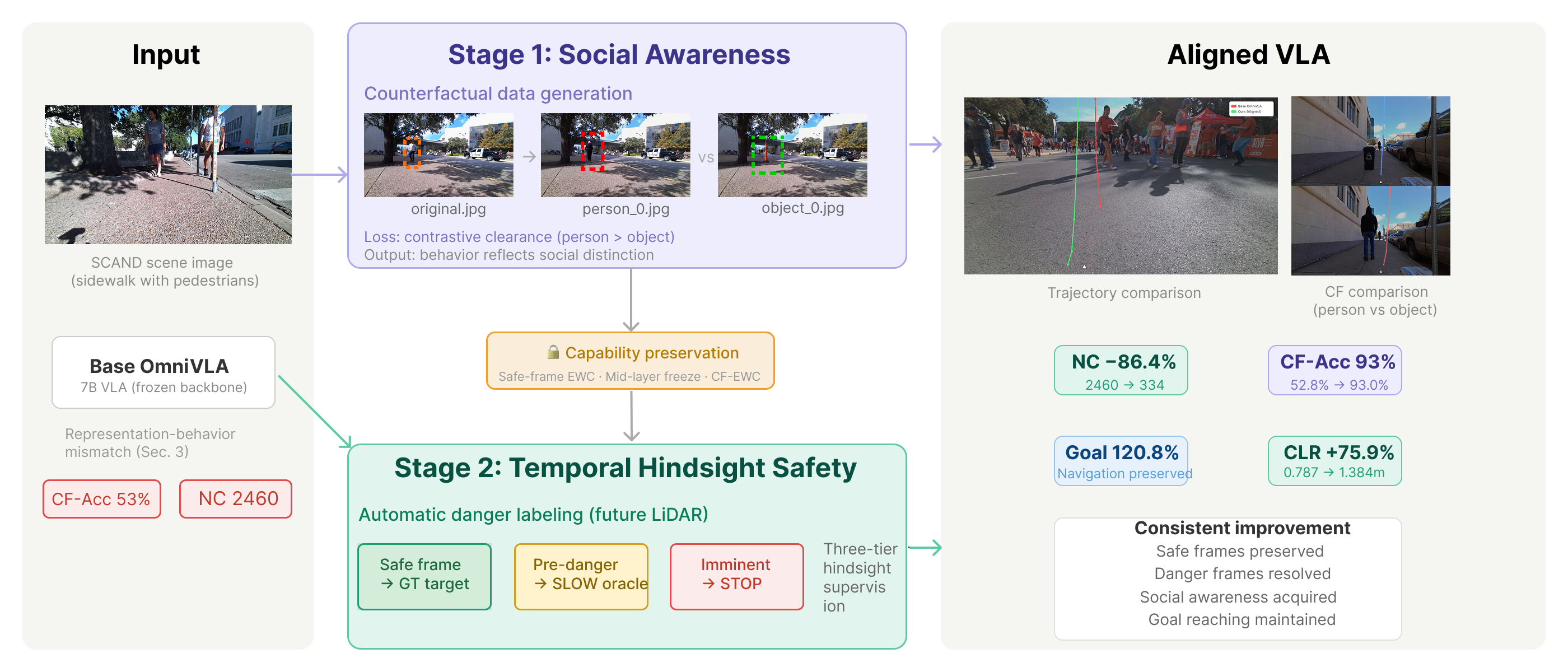}
\vspace{-15pt}
\caption{\textbf{Overview of our two-stage alignment framework.} Social Behavioral Alignment trains social awareness through counterfactual contrastive learning. Temporal Safety Alignment teaches anticipatory safety via temporal hindsight relabeling. Capability preservation (EWC) ensures both capabilities coexist.}
\vspace{-10pt}
\label{fig:method}
\end{figure*}



The analysis above reveals two perception-action gaps: both social and temporal danger signals are encoded but not acted upon, the social signal most decodable at intermediate layers and the danger signal uniformly across depth.
We therefore propose \ours{}, a two-stage post-training framework that turns these latent signals into action-level supervision without human social annotation or extra inference modules. 
The first stage addresses social distinction by exposing the action head to mid-layer social features and training on counterfactual person--object pairs. 
The second stage addresses temporal anticipation by relabeling states with future collision outcomes and training graduated safety responses. 
Cross-stage protection preserves the social behavior learned in the first stage during temporal fine-tuning. Figure~\ref{fig:method} illustrates the framework. All training uses LoRA~\cite{hu2022lora} with frozen base model parameters.

\subsection{Social Behavioral Alignment}
\label{sec:stage1}

As shown in Section~\ref{sec:representation_analysis}, social decodability peaks at intermediate layers. Social Behavioral Alignment therefore intervenes at both the representation and training levels: we give the action head direct access to these more salient intermediate-layer features, and train it on counterfactual scene pairs that provide behavioral gradients for social differentiation. To construct these pairs, we use detection, segmentation, and inpainting to replace pedestrians in SCAND frames with static objects at the same location, yielding 26,663 person--object pairs. We also generate 15,236 appearance-variant pairs to reduce spurious dependence on specific pedestrian appearances. Details of the construction pipeline are provided in Appendix~\ref{app:implementation}.

\textbf{Mid-Layer Feature Fusion. } 
The original action head reads only the final-layer representation, where the social distinction is less salient than at intermediate layers (Section~\ref{sec:representation_analysis}). We extend it to additionally receive an intermediate-layer representation near the salience peak (Layer~9), giving the head direct access to the more separable social signal.

\textbf{Counterfactual Contrastive Training. }
Feature fusion gives the action head access to social information, but it does not by itself ensure that this information affects behavior. We therefore train on matched person--object pairs and require the predicted trajectories to reflect the semantic difference between the two scenes. The objective is:
$
    \mathcal{L}_{S1} = \mathcal{L}_{bc} + \lambda_{cf}\,\mathcal{L}_{cf} + \lambda_{sem}\,\mathcal{L}_{sem} + \lambda_{inv}\,\mathcal{L}_{inv},
$
where $\mathcal{L}_{bc} = \text{MSE}(\hat{\tau}, \tau_{GT})$ anchors the policy to the original demonstrations and prevents behavioral drift. The counterfactual term $\mathcal{L}_{cf} = \max(0, m - (c_{\text{person}} - c_{\text{object}}))$, with $m = 0.15$\,m, encourages larger clearance around people than around matched static objects and vanishes once the margin is satisfied. The semantic term $\mathcal{L}_{sem}$ is a cross-entropy loss on the action-token predictions that preserves the backbone's pretrained reasoning. The invariance term $\mathcal{L}_{inv} = \text{MSE}(\hat{\tau}_{app1}, \hat{\tau}_{app2})$ penalizes trajectory differences across appearance-varied versions of the same pedestrian scene, discouraging reliance on clothing, pose, age, or gender. We further apply advantage-conditioned regression with labels $+1$ for person scenes and $-1$ for object scenes, encouraging conservative avoidance near people while discouraging unnecessary conservatism near objects.

\subsection{Temporal Safety Alignment}
\label{sec:stage2}
 

Behavioral cloning provides frame-by-frame ``observation $\rightarrow$ action'' supervision, but it does not tell the model which current states lead to unsafe future outcomes. Temporal Safety Alignment adds this missing feedback by using hindsight relabeling to connect present observations with near-future collision risk.
 

\textbf{Hindsight Temporal Labeling. }
Since SCAND is continuously recorded, future outcomes can be used to retrospectively label each frame. For each frame, we compute the minimum obstacle clearance over the next 3 frames ($\sim$1\,s at 3\,Hz). If this future clearance falls below 0.3m, the current frame is labeled as pre-danger, indicating that preventive action should already begin.


\textbf{Three-Tier Danger Response.}
Instead of using binary safe/danger targets, we train a graduated response that mirrors human navigation behavior: NORMAL for future clearance above $0.3$\,m, trained on ground-truth trajectories; SLOW+AVOID for $0.15$--$0.3$\,m, trained on oracle evasion trajectories scaled by $0.3$; and STOP for clearance below $0.15$\,m, trained on zero trajectories. This design teaches different response intensities and introduces stopping behavior, which is nearly absent from SCAND demonstrations. Danger frames receive 5--10$\times$ training weight to address class imbalance. Full details of the tier design and oracle trajectory construction are provided in Appendix~\ref{app:implementation}.

\subsection{Capability Protection Across Stages}
\label{sec:protection}

Temporal Safety Alignment fine-tunes shared LoRA parameters and action-head weights on risk-focused targets, which can overwrite the social behavior learned through Social Behavioral Alignment. To reduce this interference, we regularize the remaining trainable parameters with Elastic Weight Consolidation (EWC)~\cite{kirkpatrick2017ewc}. The Temporal Safety Alignment loss is 
$
    \mathcal{L}_{S2} = \mathcal{L}_{\text{hindsight}} + \mathcal{L}_{\text{EWC}},
$
where $\mathcal{L}_{\text{hindsight}}$ trains the model to decelerate or stop based on future collision outcomes (Section~\ref{sec:stage2}). The EWC term, $\mathcal{L}_{\text{EWC}} = \frac{\lambda_{\text{EWC}}}{2} \sum_i F_i^{\text{safe}} (\theta_i - \theta_i^*)^2$, penalizes deviation from the Social Behavioral Alignment parameters $\theta_i^*$, weighted by the diagonal Fisher Information Matrix $F_i^{\text{safe}} = \mathbb{E}_{\text{safe}}\left[\left(\frac{\partial \mathcal{L}}{\partial \theta_i}\right)^2\right]$ computed over safe frames. Parameters with high Fisher values are treated as important for safe-frame behavior and are therefore protected most strongly during Temporal Safety Alignment.

\section{Experimental Results}
\label{sec:result}


We evaluate on SCAND dataset~\cite{karnan2022socially}, an outdoor sidewalk
navigation dataset with synchronized RGB, LiDAR, and odometry at
${\sim}3$\,Hz. We use 13 held-out bags (4{,}902 frames) for evaluation
with zero scene overlap with training data. Our base model is
OmniVLA~\cite{hirose2025omnivla} (7B parameters), fine-tuned via LoRA with the
backbone frozen. We additionally evaluate three vision-based navigation
policies, GNM~\cite{shah2023gnm}, ViNT~\cite{shah2023vint}, and NoMaD~\cite{sridhar2024nomad},
on the same split under identical conditions.

We evaluate along three dimensions.
For \emph{safety}, we report Near-Collision Count (NC, frames with minimum obstacle clearance $<0.6$\,m) and Average Minimum Obstacle Clearance.
For \emph{social awareness}, we report Counterfactual Accuracy (CF-Acc)
and Differential Clearance over person--object scene pairs.
For \emph{navigational capability}, we report Goal Completion Ratio.
Implementation details are in Appendix~\ref{app:implementation}.

\begin{table}[t]
\centering
\caption{\textbf{Main results on SCAND evaluation set}
(4{,}902 frames). }
\label{tab:main}
\resizebox{0.6\columnwidth}{!}{
\begin{tabular}{l ccccc}
\toprule
    \textbf{Method}
    & \textbf{NC} $\downarrow$
    & $\Delta$\textbf{NC}
    & \textbf{Clr.\ (m)} $\uparrow$
    & \textbf{CF-Acc} $\uparrow$
    & \textbf{Goal} $\uparrow$ \\
\midrule
    GNM       & 1{,}906 & ---          & 1.106 & 31   & 8.1\%  \\
    ViNT      & 1{,}702 & ---          & 1.146 & 31   & 8.8\%  \\
    NoMaD     & 511     & ---          & 1.644 & 44   & 0.4\% \\
    OmniVLA   & 2{,}460 & ---          & 0.787 & 52.8 & 108.9\% \\
\midrule
    Ours (S1) & 2{,}580 & $+$4.9\%    & 0.675 & \textbf{96.0} & 85.3\% \\
    \textbf{Ours}
              & \textbf{334} & $\mathbf{-86.4\%}$
              & \textbf{1.384} & 93.0 & 96.7\% \\
\bottomrule
\end{tabular}
}
\vspace{-12pt}
\end{table}

\textbf{Social Awareness.} Table~\ref{tab:main} shows that the base OmniVLA achieves only 52.8\% counterfactual accuracy and near-zero differential clearance (+0.002\,m), indicating little behavioral distinction between pedestrians and matched static objects. Our framework increases CF-Acc to 93.0\% and differential clearance to +0.045\,m, showing that the policy learns to give pedestrians more space than non-social obstacles with the same geometry.

\begin{figure}[t]
\centering
\begin{subfigure}[t]{0.32\linewidth}
  \centering
  \includegraphics[width=\linewidth]{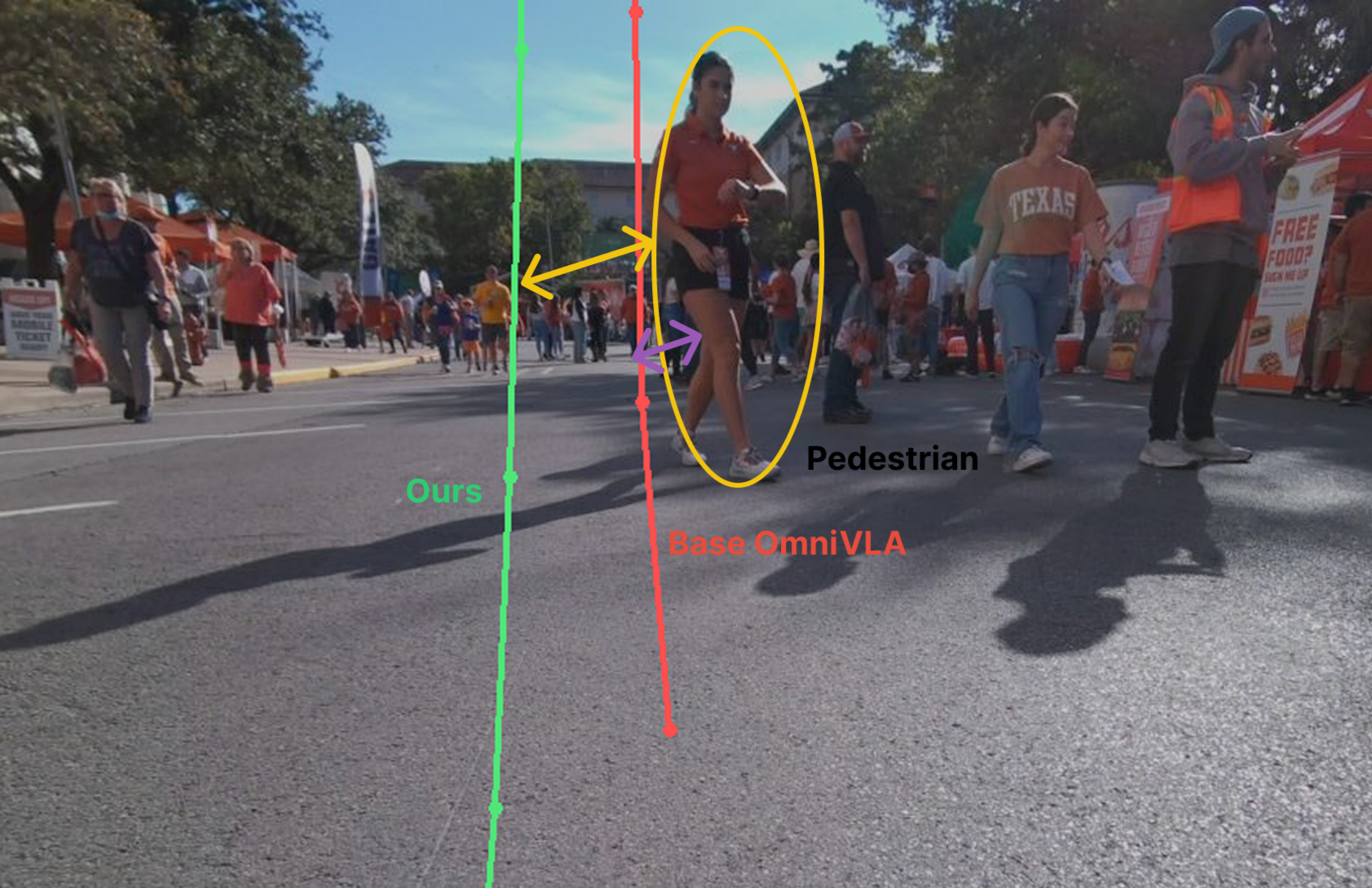}
  \caption{Proactive safety}
  \label{fig:proactive_safety}
\end{subfigure}
\hfill
\begin{subfigure}[t]{0.32\linewidth}
  \centering
  \includegraphics[width=\linewidth]{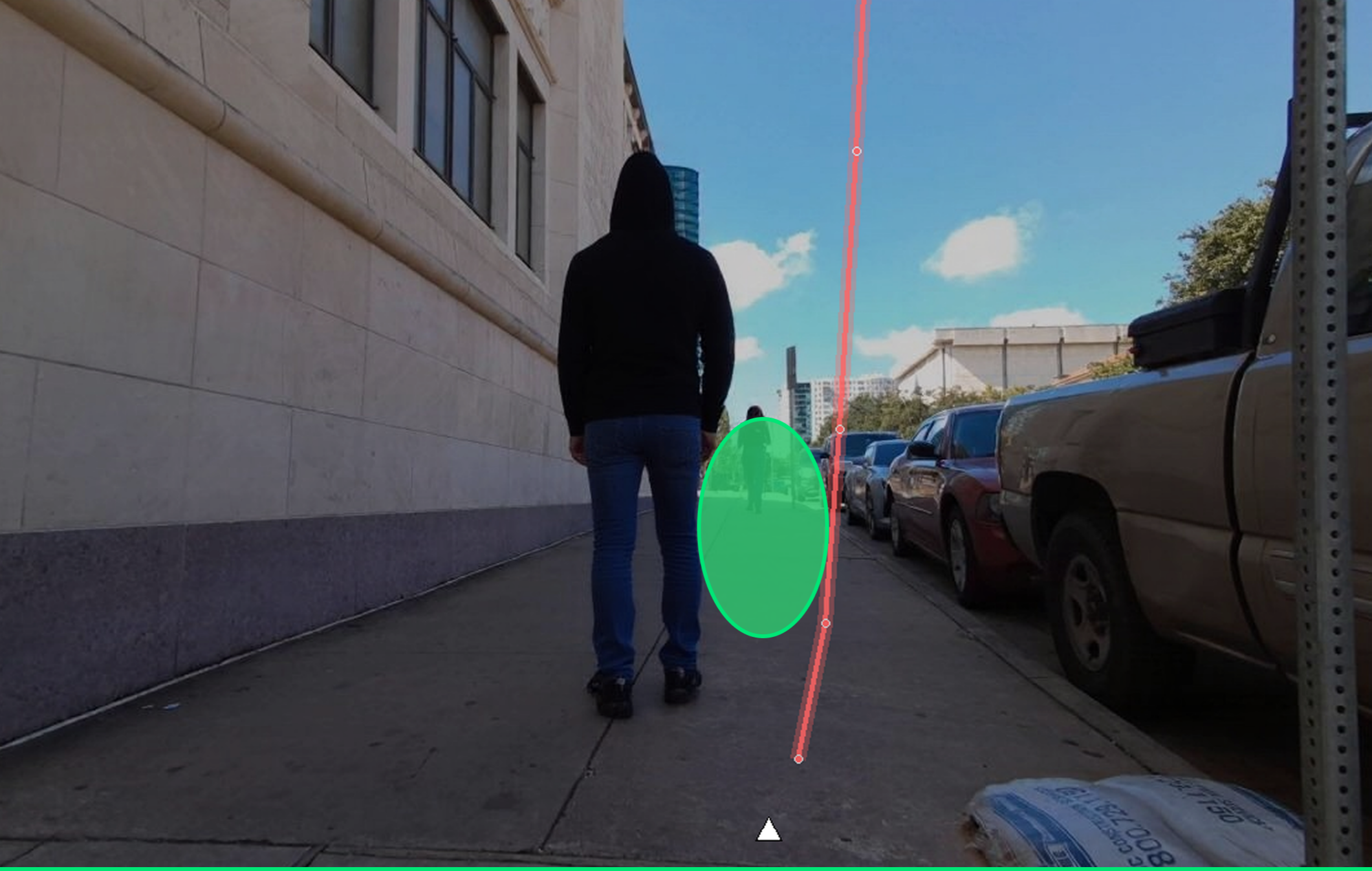}
  \caption{Person present}
  \label{fig:cf_person}
\end{subfigure}
\hfill
\begin{subfigure}[t]{0.32\linewidth}
  \centering
  \includegraphics[width=\linewidth]{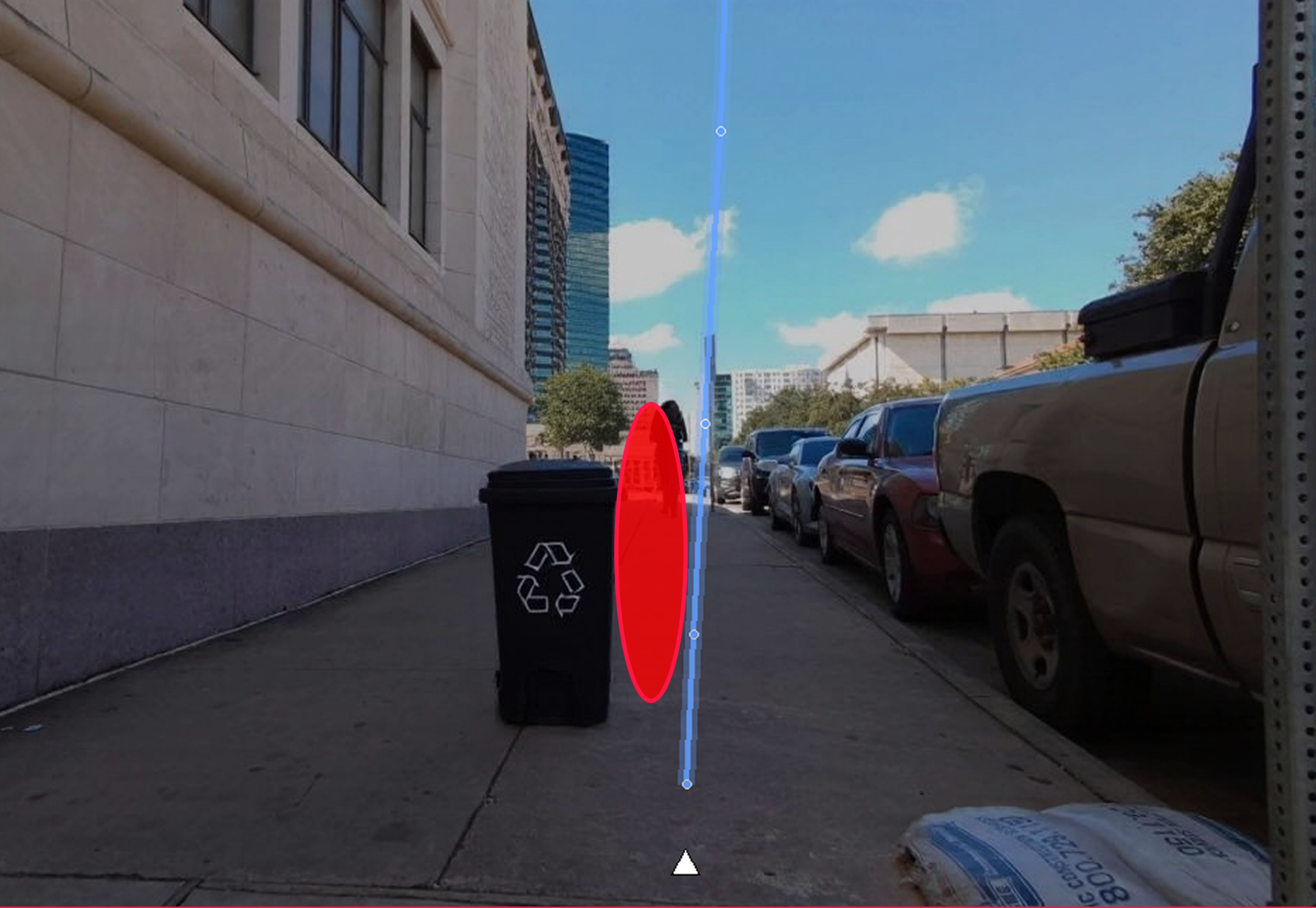}
  \caption{Object replaced}
  \label{fig:cf_object}
\end{subfigure}

\caption{
\textbf{Qualitative comparison of safety and social awareness.}
(a) The base model's trajectory leads to a near-collision, whereas our model anticipates the risk and adjusts its course early.
(b,c) In the same scene, replacing a person with an object changes the predicted behavior: our model gives the person greater clearance, indicating sensitivity to socially relevant scene semantics.
}
\label{fig:safety_social_awareness}
\vspace{-12pt}
\end{figure}

\textbf{Safety.}
\textbf{\textit{1) Main results.}} Our method reduces near-collisions by 86.4\% (from 2{,}460 to 334) and increases average obstacle clearance by 75.9\%. GNM and ViNT exhibit lower NC counts than OmniVLA but achieve only 8\% goal completion, indicating near-stationary rather than safer trajectories.
%
%
\textbf{\textit{2) Safe vs.\ danger frame analysis.}}  We split evaluation frames by ground-truth future clearance. OmniVLA reaches only 0.305\,m mean clearance on danger frames, indicating little anticipatory avoidance, while our model increases danger-frame clearance to 1.251\,m (+310\%) and slightly improves safe-frame clearance (1.273\,m to 1.520\,m). Table~\ref{tab:frame_analysis}(Appendix) shows that our method resolves 90.0\% of danger frames with only 3.6\% safe-frame degradation; without EWC, this degradation rises to 25\%.
%
%
\textbf{\textit{3) Goal reaching.}} In sequential open-loop evaluation, our model achieves a completion ratio of 96.7\% relative to human demonstrations, 0\% stop rate, and zero stall segments across all evaluation sequences, confirming that improved safety does not compromise continuous, 
goal-directed navigation.

\textbf{Ablation Study.}
Table~\ref{tab:ablation} isolates three design choices.
\textbf{\textit{1) Component contribution.}} Social behavioral alignment improves CF-Acc from 52.8\% to 96.0\% but does not improve safety, while temporal safety alignment alone reduces NC by 57.6\% but leaves CF-Acc at 36.0\%. Only the full pipeline achieves both, reducing NC by 86.4\% while maintaining 93.0\% CF-Acc.
\textbf{\textit{2) Capability preservation.}} Mid-layer projection is the main mechanism for preserving social awareness during temporal safety alignment. Without it (B0, standard LoRA), NC is still reduced, but CF-Acc collapses to 41\%, suggesting that social and safety objectives interfere when they share the same parameter pathway. Adding the projection creates a separate channel for social information, recovering CF-Acc to 74\% even without explicit protection (B1). Safe-frame EWC (B2) further improves both NC reduction and CF-Acc, mid-layer freezing (B3) strengthens safety with a small CF-Acc trade-off, and counterfactual EWC (B4) restores CF-Acc to 93\% with negligible loss in NC reduction.
\textbf{\textit{3) Temporal supervision design.}} Without temporal supervision (C1), NC reduction reaches only 45.7\%, indicating that behavior cloning alone provides limited anticipatory safety. Binary danger labels (C2) improve this to 70.2\% by distinguishing safe and dangerous frames, but still encourage relatively late reactions. The three-tier design (C3) reaches 86.4\% by adding a pre-danger tier, which teaches the policy to decelerate before collisions become imminent rather than stopping at the last moment. Similar CF-Acc across all three variants shows that temporal supervision improves safety without disrupting social awareness.

\begin{table*}[t]
\begin{center}
\resizebox{0.8\textwidth}{!}{
\begin{tabular}{cl cccc}
\toprule
    & \textbf{Configuration}
    & \textbf{NC} $\downarrow$
    & $\Delta$\textbf{NC}
    & \textbf{CF-Acc (\%)} $\uparrow$
    & \textbf{Clearance (m)} $\uparrow$ \\
\midrule
\multicolumn{6}{l}{\textit{Group A: Stage Contribution}} \\
\midrule
    A1 & Base OmniVLA                   & 2460 & ---         & 52.8 & 0.787 \\
    A2 & Stage~1 only                   & 2580 & $+$4.9\%   & \textbf{96.0} & 0.675 \\
    A3 & Stage~2 only (no S1)           & 1043   & $-$57.6\%      & 36.0 & 1.12 \\
    A4 & Stage~1 $+$ Stage~2            & \textbf{334} & $\mathbf{-86.4\%}$ & 93.0 & \textbf{1.384} \\
\midrule
\multicolumn{6}{l}{\textit{Group B: Capability Preservation}} \\
\midrule
    B0 & Standard LoRA (no mid-proj)   & 326    & $-$86.7\%  & 41  & 1.405 \\
    B1 & No protection                  & 1593   & $-$35.2\%      & 74 & 1.042 \\
    B2 & $+$ Safe-frame EWC             & 671   & $-$72.7\%      & 87 & 1.291 \\
    B3 & $+$ Mid-layer freeze           & 374   & $-$84.8\%      & 84 & 1.380 \\
    B4 & $+$ Counterfactual EWC         & 334     & $-$86.4\%  & \textbf{93} & \textbf{1.384} \\
\midrule
\multicolumn{6}{l}{\textit{Group C: Temporal Supervision Design}} \\
\midrule
    C1 & No temporal supervision         & 1336   & $-$45.7\%      & 94 & 1.060 \\
    C2 & Binary danger labels            & 734   & $-$70.2\%      & 95 & 1.271 \\
    C3 & Three-tier response targets     & \textbf{334} & $\mathbf{-86.4\%}$ & 93 & \textbf{1.384} \\
\bottomrule
\end{tabular}
}
\end{center}
\vspace{-5pt}
\caption{\textbf{Comprehensive ablation study.}\textbf{Group~A} isolates stage contributions: A1 is the unmodified base model; A2 applies only social alignment (Stage~1); A3 applies only temporal safety (Stage~2) without Stage~1; A4 is the full two-stage pipeline.  \textbf{Group~B} tests capability preservation mechanisms during Stage~2, starting from no mid-layer projection (B0), then progressively adding: mid-layer projection without protection (B1), safe-frame EWC (B2), mid-layer freezing (B3), and counterfactual EWC (B4). Each layer provides cumulative benefit. \textbf{Group~C} compares temporal supervision designs: no temporal relabeling (C1), binary safe/dangerous labels (C2), and our three-tier safe/pre-danger/danger targets (C3). The three-tier design achieves the best NC reduction (\textbf{bold}) while all variants maintain similar CF-Acc, confirming that temporal supervision does not interfere with social awareness.}
\label{tab:ablation}
\vspace{-10pt}
\end{table*}

 \begin{figure}[t]
    \centering
    \includegraphics[width=\linewidth]{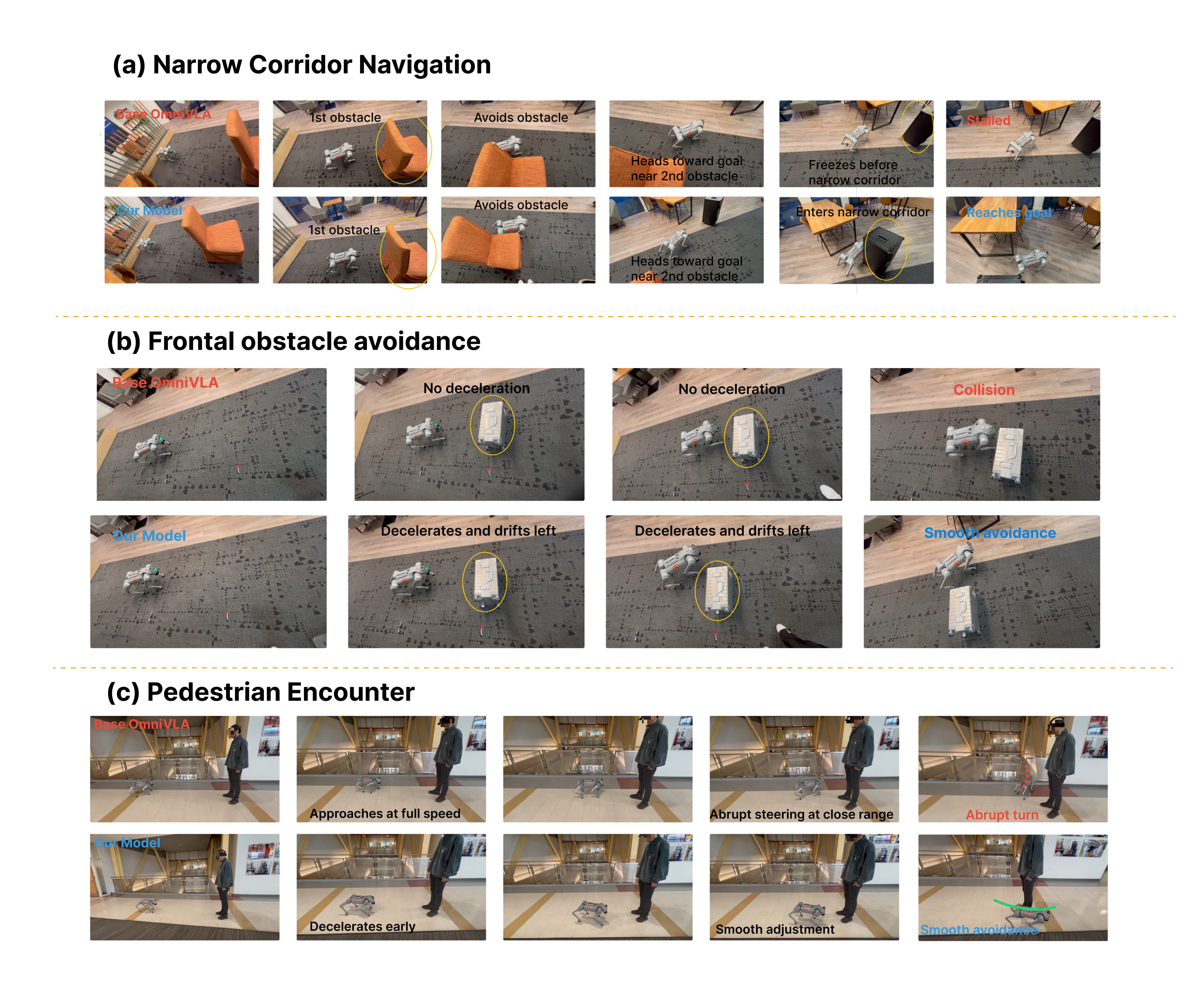}
    \vspace{-30pt}
    \caption{\textbf{Real-world deployment on a Unitree quadruped.} Each scenario shows the base OmniVLA (top row, red label) and our aligned model (bottom row, blue label) across key timesteps. \textbf{(a)} Narrow corridor navigation: both models avoid the first obstacle, but the base model freezes before the narrow corridor formed by the second obstacle, while ours navigates through and reaches the goal. \textbf{(b)} Frontal obstacle avoidance: the base model shows no deceleration and collides at step 10, while ours begins decelerating at steps 6--7 and smoothly avoids the obstacle. \textbf{(c)} Pedestrian encounter: the base model holds its heading until close range and then makes an abrupt avoidance maneuver, while ours begins adjusting earlier and steers smoothly around the pedestrian.}
    \label{fig:real_world}
    \vspace{-10pt}
\end{figure}
\subsection{Real-World Deployment}
\label{sec:q4}

We evaluate real-world transfer by deploying the aligned OmniVLA model on a Unitree quadruped with LoRA adapters and monocular RGB only; no LiDAR is used at inference. We compare against the base VLA in three scenarios that stress complementary capabilities: constrained goal-directed navigation, sudden obstacle response, and human-aware navigation.

\textbf{1) Narrow Corridor Navigation.} Two obstacles are placed along the robot's route. Both models avoid the first obstacle, but the second forms a narrow corridor near the goal, requiring the policy to continue making goal-directed progress under tight spatial constraints. The base VLA stops before entering the corridor, whereas our model passes through it and reaches the goal. \textbf{2) Frontal Obstacle Avoidance.} An obstacle is suddenly placed directly in the robot's path during forward motion. The base VLA drifts left without slowing down and collides at step 10. In contrast, our model begins decelerating at steps 6--7, then steers smoothly around the obstacle and continues toward the goal. 
\textbf{3) Pedestrian Encounter.}A person approaches and stops in front of the robot. The base VLA holds its heading until close range and then makes an abrupt avoidance maneuver, whereas our model begins adjusting earlier and follows a smoother, wider path around the pedestrian. Across all three scenarios, the aligned model transfers to the real robot with earlier anticipatory adjustments while preserving goal-directed, socially smoother navigation.

\subsection{Discussion}
\label{sec:q5}

\textbf{Perception--Action Mismatch as a Diagnostic Framework.}
Comparing internal representations with behavior provides a practical way to localize bottlenecks: architectural interventions address limited access to encoded information, while improved supervision addresses failures to act on it.

\textbf{Breaking the Saliency Shortcut.}
The base model relies on a shortcut that avoids visually salient regions rather than acting on social context. Stage~1 counterfactual training breaks this shortcut by forcing the model to differentiate based on what an obstacle is, not how prominent it looks.

\textbf{From Reactive to Anticipatory Safety.}
Temporal hindsight relabeling provides the missing causal signal linking current observations to future consequences, shifting the model from reactive to anticipatory decision-making.

\vspace{-6pt}
\section{Conclusion}
\label{sec:conclusion}


In this paper, we identify a perception--action mismatch in navigation VLAs: pretrained representations encode pedestrian--object distinctions and future collision risk, but the policy does not reliably express these signals in behavior. \ours{} addresses this gap through counterfactual human--object alignment and temporal hindsight relabeling, improving social differentiation and anticipatory safety on SCAND and in real-world deployment. These results show that safer social navigation can be obtained by aligning VLA actions with existing internal representations, without human social annotations or additional inference-time reasoning modules.

\textbf{Limitations.} Our primary evaluation is offline on SCAND, using LiDAR-computed distance metrics that may not capture the full complexity of closed-loop interaction. Although we include real-world deployment, broader validation across denser pedestrian crowds, diverse platforms, and longer-horizon navigation tasks remains necessary. In addition, our social metrics focus mainly on clearance-based distinctions and do not yet evaluate richer social behaviors such as intent anticipation, yielding, or negotiation.



\clearpage
\acknowledgments{If a paper is accepted, the final camera-ready version will (and probably should) include acknowledgments. All acknowledgments go at the end of the paper, including thanks to reviewers who gave useful comments, to colleagues who contributed to the ideas, and to funding agencies and corporate sponsors that provided financial support.}


\bibliography{example}  


\newpage
\appendix

\section{Detailed Real-Robot Deployment Results}
%

\begin{table*}[h]
\centering
\caption{Detailed real-robot deployment metrics per scenario, computed over the
logged trials with full telemetry. 
\textbf{Goal reached} is counted if the robot ends within 0.5\,m of the goal.
\textbf{Steps} is the mean trajectory length (a
collision trial ends at the collision step; a non-reaching trial ends at the 25-step
budget). \textbf{Smoothness} is the mean total variation of the commanded yaw-rate
over non-collision trials (lower is smoother). \textbf{Final Goal Dist.} is the mean
end-of-trial distance to the goal over non-collision trials (lower is closer). The
Sequential scenario uses a 5\,m goal and a 30-step budget and is reported separately.}
\label{tab:realrobot-detailed}
\small
\setlength{\tabcolsep}{5pt}
\begin{tabular}{l l c c c c c c}
\toprule
 &  &  & Goal Reached & Collision &  & Smoothness & Final Goal \\
Scenario & Method & $n$ & (\%) $\uparrow$ & (\%) $\downarrow$ & Steps & (vyaw TV) $\downarrow$ & Dist.\ (m) $\downarrow$ \\
\midrule
Static            & \textbf{SALSA (ours)} & 5  & \textbf{80}  & \textbf{0}  & 21.6 & \textbf{0.277} & \textbf{0.43} \\
                  & OmniVLA               & 5  & 60           & 40          & 17.2 & 0.400          & 1.45 \\
\addlinespace
Narrow            & \textbf{SALSA (ours)} & 18 & \textbf{100} & \textbf{0}  & 20.6 & \textbf{0.287} & \textbf{0.33} \\
                  & OmniVLA               & 13 & 15           & 31          & 19.2 & 0.342          & 1.04 \\
\addlinespace
Pedestrian        & \textbf{SALSA (ours)} & 5  & \textbf{100} & \textbf{0}  & 20.5 & \textbf{0.328} & \textbf{0.27} \\
                  & OmniVLA               & 5  & 20            & 20          & 20.8 & 0.554          & 1.56 \\
\addlinespace
Dynamic           & \textbf{SALSA (ours)} & 8  & \textbf{100} & \textbf{0}  & 21.0 & \textbf{0.345} & \textbf{0.25} \\
                  & OmniVLA               & 8  & 0            & 25          & 22.2 & 0.541          & 1.44 \\
\midrule
Sequential (5\,m) & \textbf{SALSA (ours)} & 2  & \textbf{100} & \textbf{0}   & 26.0 & 0.485 & 0.21 \\
                  & OmniVLA               & 2  & 0            & 100          & 15.0 & n/a   & n/a \\
\bottomrule
\end{tabular}
\end{table*}

\section{Implementation Details}
\label{app:implementation}

\subsection{Three-Tier Danger Response}
 
A binary safe/danger label is insufficient to guide nuanced behavior. Human navigators exhibit graduated responses: proceeding normally at long range, decelerating at medium range, and stopping at close range. We design a corresponding three-tier mechanism:
 
\begin{table}[h]
\centering
\caption{Three-tier danger response mechanism. The oracle trajectory is the candidate with the highest clearance among 25 candidates per frame, scaled by 0.3 to enforce simultaneous deceleration during evasion.}
\label{tab:threetier}
\begin{tabular}{lll}
\toprule
Future Min Clearance & Label & Training Target \\
\midrule
$< 0.15$\,m (imminent collision) & STOP & Zero trajectory \\
$0.15$--$0.3$\,m (approaching danger) & SLOW+AVOID & Oracle traj $\times 0.3$ \\
$> 0.3$\,m (safe) & NORMAL & Ground-truth trajectory \\
\bottomrule
\end{tabular}
\end{table}

\paragraph{Clearance Computation.}
For a predicted trajectory $\hat{\tau} = \{(\Delta x_i, \Delta y_i, \cos\Delta\psi_i, \sin\Delta\psi_i)\}_{i=1}^{8}$, we first accumulate relative waypoints into absolute positions $\mathbf{w}_i$ in the robot's local frame. Given a synchronized LiDAR scan $\mathcal{P} = \{\mathbf{p}_j\}_{j=1}^{M}$ projected onto the ground plane, the clearance is defined as:
\begin{equation}
    c(\hat{\tau}, \mathcal{P}) = \min_{i \in \{1,\dots,8\}} \min_{\mathbf{p}_j \in \mathcal{P}} \|\mathbf{w}_i - \mathbf{p}_j\|_2.
\end{equation}
A frame is classified as a \textit{near-collision} if $c(\hat{\tau}, \mathcal{P}) < d_{\text{robot}} = 0.6$\,m, matching the robot's physical width.

\paragraph{Oracle Trajectory Construction.}
For each training frame, we generate a set of 25 candidate trajectories $\{\tau^{(k)}\}_{k=0}^{24}$. Candidate $\tau^{(0)}$ is the ground-truth trajectory recorded by the human operator. Candidates $\tau^{(1)}$--$\tau^{(24)}$ are constructed by rotating $\tau^{(0)}$ at uniformly spaced angular offsets, producing spatially diverse alternatives that remain dynamically plausible (preserving the original displacement magnitude). Each candidate is evaluated against the LiDAR scan:
\begin{equation}
    k^* = \arg\max_{k \in \{0,\dots,24\}} \; c(\tau^{(k)}, \mathcal{P}),
\end{equation}
and the oracle trajectory $\tau^{\text{oracle}} = \tau^{(k^*)}$ is selected as the candidate with the highest obstacle clearance. For SLOW+AVOID frames, the training target is a speed-scaled version of the oracle: $\tau^{\text{target}} = 0.3 \times \tau^{\text{oracle}}$, reducing displacement to 30\% while preserving the oracle's lateral avoidance direction. For STOP frames, the target is a stationary trajectory $\tau^{\text{target}} = [0, 0, 1, 0]^{\times 8}$.

This design reflects three considerations. Graduated targets avoid the pathologies of binary labeling, allowing the model to learn different response intensities rather than either global deceleration or reaction only to extreme danger. The STOP target explicitly introduces stopping behavior, which is nearly absent from SCAND's demonstrations yet critical in real deployment. Finally, danger frames receive 5–10$\times$training weight to ensure sufficient learning despite severe class imbalance.

\subsection{Counterfactual Benchmark Construction}
\label{sec:cfgen}
 
To construct training data for Stage 1, we design an automated pipeline that alters only the semantic identity of entities while preserving the geometric structure of the scene.
 
\textbf{Detection and Filtering.}
For each pedestrian-containing frame in SCAND, we run YOLOv8 for detection and SAM2 for pixel-precise segmentation. We then filter by area ratio, image position, and spatial relation to the robot's future trajectory, retaining only pedestrians with meaningful impact on the current navigation decision.
 
\textbf{Semantic Counterfactual Generation.}
Using FLUX.1 Fill Dev, we inpaint each masked region with one of seven static object categories: trash can, traffic cone, fire hydrant, potted plant, stone bollard, yellow parking bollard, and recycling bin. Using text prompts to constrain the target category and scene context, while background geometry, lighting, and viewpoint are preserved. For scenes with 3 or more people, all pedestrians are replaced simultaneously to produce stronger semantic contrast; for 1--2 people, pedestrians are replaced individually with randomized object types.
 
\textbf{Appearance Invariance Augmentation.}
To prevent the model from learning spurious correlations with specific pedestrian appearances, We use the same pipeline to vary the pedestrian's clothing, pose, age, and gender without changing the scene content. This forces the model to attend to pedestrian presence rather than specific visual patterns.
 
\textbf{Data Scale.}
The pipeline identifies 3,809 valid frames, yielding approximately 26,663 person-object pairs and 15,236 appearance-variant pairs.

\textbf{Training Objective.}
Stage~1 optimizes a composite loss:
\begin{equation}
    \mathcal{L}_{\text{S1}} = \mathcal{L}_{\text{bc}} + \lambda_{\text{cf}}\,\mathcal{L}_{\text{cf}} + \lambda_{\text{sem}}\,\mathcal{L}_{\text{sem}} + \lambda_{\text{inv}}\,\mathcal{L}_{\text{inv}},
\end{equation}
where each term serves a distinct role:
\begin{itemize}
    \item $\mathcal{L}_{\text{bc}} = \text{MSE}(\hat{\tau},\, \tau_{\text{GT}})$ anchors predictions to ground-truth trajectories, preventing output drift;
    \item $\mathcal{L}_{\text{cf}} = \max\!\big(0,\; m - (c_{\text{person}} - c_{\text{object}})\big)$ is a hinge loss with margin $m = 0.15$\,m that encourages greater clearance for pedestrian scenes than for matched object scenes;
    \item $\mathcal{L}_{\text{sem}}$ is a cross-entropy loss on the action token predictions, preserving the language backbone's semantic reasoning capacity;
    \item $\mathcal{L}_{\text{inv}}$ penalizes trajectory variance across appearance-augmented versions of the same pedestrian scene, enforcing invariance to visual appearance.
\end{itemize}
We set $\lambda_{\text{cf}} = 1.0$, $\lambda_{\text{sem}} = 3.0$, and $\lambda_{\text{inv}} = 0.3$.

\section{Additional Results}
\label{app:results}

\begin{table}[h]
\centering
\caption{\textbf{Safe vs.\ danger frame analysis.} A practical safety method must resolve danger frames without degrading safe ones. Originally dangerous: 2{,}460 frames; originally safe: 2{,}442 frames.}
\label{tab:frame_analysis}
\begin{tabular}{l ccc}
\toprule
\textbf{Method}
    & \textbf{Danger Resolved} $\uparrow$
    & \textbf{New NC on Safe} $\downarrow$
    & \textbf{Total NC} $\downarrow$ \\
\midrule
OmniVLA (base)
    & 0\%
    & 0\%
    & 2{,}460 \\
GNM
    & 44.1\%
    & 21.7\%
    & 1{,}906 \\
ViNT
    & 51.5\%
    & 20.8\%
    & 1{,}702 \\
NoMaD
    & 84.1\%
    & 4.9\%
    & 511 \\
\midrule
Ours (w/o EWC)
    & 60.4\%
    & 25.2\%
    &  1{,}590\\
\textbf{Ours (Full)}
    & \textbf{90.0\%}
    & \textbf{3.6\%}
    & \textbf{334} \\
\bottomrule
\end{tabular}
\end{table}

\end{document}